\pdfoutput=1

\documentclass[11pt]{article}

\usepackage[]{acl}

\usepackage{times}
\usepackage{latexsym}
\usepackage{xcolor,colortbl}
\usepackage[T1]{fontenc}
\usepackage{pifont}
\usepackage[utf8]{inputenc}
\usepackage{microtype}
\usepackage{amsmath}
\usepackage{amssymb}

\usepackage{inconsolata}
\usepackage{graphicx}
\usepackage{booktabs}
\usepackage{mathtools}

\newcommand{\our}{\texttt{Ask, Refine, and Trust}}
\newcommand{\art}{\texttt{ART}}
\newcommand{\ourfull}{\texttt{ART: Ask, Refine, and Trust}}
\newcommand{\asker}{\texttt{Asker}}
\newcommand{\truster}{\texttt{Truster}}

\definecolor{Yellow}{rgb}{1,1,0.8}
\definecolor{Blue}{rgb}{0.6,0.8,0.9}
\definecolor{Green}{rgb}{0.75,0.87,0.75}

\title{The \art\ of LLM Refinement: \our}

\author{%
  Kumar Shridhar $^{\diamond}$ \thanks{\quad Work done during internship at Meta AI; correspondence at \texttt{shkumar@ethz.ch}}\ \quad Koustuv Sinha $^{\spadesuit}$ \quad  Andrew Cohen $^{\spadesuit}$ \quad  Tianlu Wang $^{\spadesuit}$ \quad Ping Yu $^{\spadesuit}$ \\ \bf{ Ram Pasunuru $^{\spadesuit}$ \quad Mrinmaya Sachan $^{\diamond}$ \quad Jason Weston $^{\spadesuit}$ \quad Asli  Celikyilmaz $^{\spadesuit}$ }\\
  \\
  $^{\diamond}$ ETH Zurich \quad $^{\spadesuit}$ Meta AI \\
  }

\begin{document}
\maketitle
\begin{abstract}

Large Language Models (LLMs) have demonstrated remarkable generative abilities, but can they judge the quality of their own generations?
A popular concept, referred to as \textit{self-refinement}, postulates that LLMs can detect and correct the errors in their generations when asked to do so. However, recent empirical evidence points in the opposite direction, suggesting that LLMs often struggle to accurately identify errors when reasoning is involved. To address this, we propose a reasoning with refinement strategy called \ourfull{}, which \emph{asks} necessary questions to decide when an LLM should \emph{refine} its output, and either affirm or withhold \emph{trust} in its refinement by ranking the refinement and the initial prediction. On two multistep reasoning tasks of mathematical word problems (GSM8K) and question answering (StrategyQA), \art{} achieves a performance gain of 5 points over self-refinement baselines, while using a much smaller model as the decision maker. 
We also demonstrate the benefit of using smaller models to make refinement decisions as a cost-effective alternative to fine-tuning a larger model.
\end{abstract}

\section{Introduction}

\begin{figure*}[t]
\includegraphics[width=\textwidth]{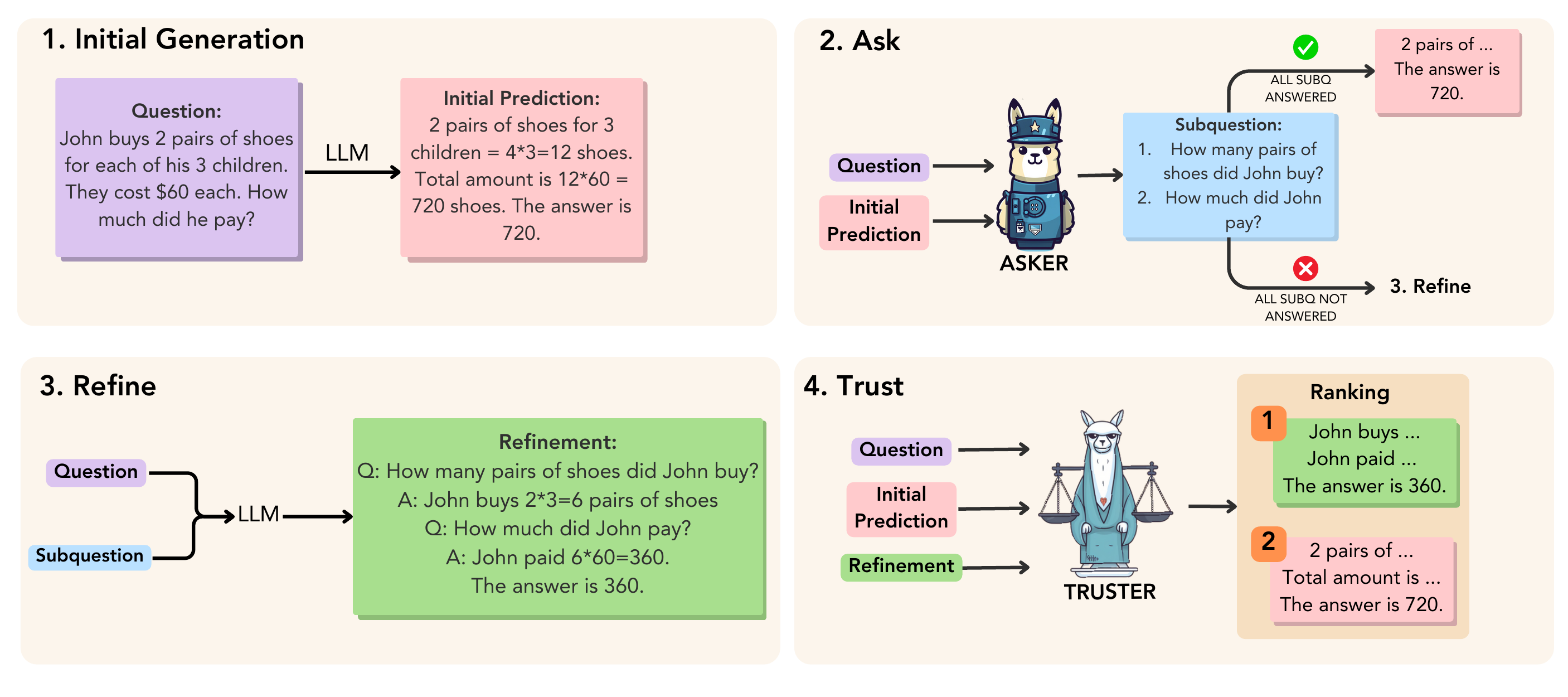}
    \caption{Our proposed objective: \ourfull{} during inference. Given a problem, an LLM first generates an initial prediction which is sent to an \asker{} that asks relevant questions (sub-questions) to decide whether refinement is needed or not. If all sub-questions are answered, it returns the initial prediction and no refinement is needed. If not, the model refines the initial prediction using the subquestions. Finally, the initial prediction and the refined  response is sent to the \truster{}, which ranks them to decide if refinement was needed or if the initial prediction was better.}
    \label{fig:main:large}
\end{figure*}

The ability of Large Language Models (LLMs) to generate coherent and meaningful text has improved significantly over the years \cite{openai2023gpt4}. However, LLMs often exhibit inaccuracies in their initial generations, and it has been posited that iterative refinement can rectify their errors \cite{SelfRefine, Shridhar2023SCREWSAM, Welleck2022GeneratingSB, php}. \citet{SelfRefine} demonstrated the potential of \textit{self-refinement}
for diverse tasks such as dialogue response and sentiment reversal; however, this approach proved less effective when applied to mathematical reasoning.
Similarly, \citet{Shridhar2023SCREWSAM} and \citet{Huang2023LargeLM} further demonstrated the challenges LLMs face in identifying errors in reasoning tasks. 
Developing models that consistently evaluate and correct their errors would be a valuable step towards building more reliable language models.

Through empirical observation on two multistep reasoning datasets, we find that \textit{self-refinement} does not reliably improve initial generations, validating the previous findings of \citet{Huang2023LargeLM}. In fact, in the majority of cases, \textit{self-refinement} has a detrimental effect on performance. 
On the other hand, fine-tuning language models usually improves their performance on a given task by facilitating better adaptation to the task objectives \cite{Yuan2023ScalingRO}. Smaller models can be trained on LLMs' data to improve their performance, which can serve as cost-effective alternatives to LLMs for the given task \cite{magister-etal-2023-teaching, shridhar-etal-2023-distilling, Hsieh2023DistillingSO}. This led us to explore the possibility of training a smaller model as a decision maker for refinement, which can consistently determine \textit{when to refine}, while the larger model can subsequently perform the refinement process.

In our work, we propose a refinement approach called \ourfull{}, which, given an initial LLM response, works in the following three stages: (a) evaluating whether the initial generation requires refinement by asking a series of questions (\texttt{Ask}); (b) executing the \textit{refinement} step based on the evaluation (\texttt{Refine}); and finally (c) selecting either the refined result or the initial prediction (\texttt{Trust}).
On two multistep reasoning tasks, mathematical reasoning and question answering, we illustrate the effectiveness of \art{} by training models of different sizes. We observe that a much smaller model \cite[LLaMA 7B;][]{LLaMA2} trained to decide \emph{when to refine}, can outperform a 10x larger model (LLaMA 70B) in a \emph{self-refinement} setup (by up to 5 points).
In addition, we evaluate the cost and accuracy tradeoffs of training a smaller model with \art{} to make a refinement decision for a pretrained LLM vs fine-tuning the LLM. In many cases, we illustrate the cost-effectiveness of \art{} as a viable alternative to fine-tuning LLMs. Finally, we show that our trained models (\asker{} and \truster{}) can work seamlessly across a wide range of  LLMs (LLaMA 70B \cite{LLaMA2}, ChatGPT \cite{brown2020language} and GPT-4 \cite{openai2023gpt4}) without requiring additional modifications.

\section{Related Work}
Strategies that use intermediate computation to solve reasoning tasks such as chain of thought \cite{cot, minerva, zhang2022automatic, kojima2022large, selfconsistency, FaithfulCot} and subquestion decomposition \cite{min2019multi,socratic_cot, leasttomost, radhakrishnan2023question} have proven to be very effective. Most LLM refinement techniques use one of these two strategies \cite{SelfRefine, Welleck2022GeneratingSB, Huang2023LargeLM, Refiner, Yoran2023AnsweringQB} or occasionally a combination of the two \cite{Shridhar2023SCREWSAM}. \citet{Shridhar2023SCREWSAM} unified past \emph{reasoning with refinement} methods under a common umbrella of \emph{sampling} (given a query, LLM generates the initial response), \emph{re-sampling} (LLM refines the initial response), and \emph{selection} (choose either the refinement or rollback to initial response). However, a single LLM was used to perform the initial generation, refinement and later selection by using different prompts. We, on the other hand, propose to train a separate, much smaller capacity model to make refinement decisions and later decide whether or not to trust the refinement over the initial generation and are not limited to prompting-based solutions. 

Asking questions to verify facts present in the model prediction has been studied by
\citet{ChainofVerification} in the context of hallucination detection. However, this work only deals with hallucinations in the form of directly stated factual inaccuracies.
It is important to note that hallucinations can take many other forms, including incorrect reasoning steps. To address this, we train an expert model to verify each reasoning step by asking relevant questions.

Training a model to rank the outputs has been studied in the past in various contexts \cite{burges}, including but not limited to text generation \cite{Krishna2022RankGenIT}, mathematical reasoning problems \cite{GSM8k}, machine translation \cite{alignment_handbook2023}, and so on. However, we do not study the standard setting of training to rank the quality of generations, but rather to decide if the refinement led to incorrect results and if it needs to be rolled back. This has some additional similarities to rejection sampling fine-tuning \cite{Yuan2023ScalingRO}, where a model is trained to generate and collect the correct reasoning chains as augmented fine-tuning datasets. On the other hand, we collect both correct and incorrect reasoning chains for ranking the outputs.

Finally, our work is similar to distilling reasoning skills into smaller models \cite{shridhar-etal-2023-distilling, magister-etal-2023-teaching, Hsieh2023DistillingSO}. However, instead of teaching smaller models to reason, we train smaller models to ask questions to verify the reasoning and decide whether the reasoning is correct, which differs from asking questions as planning to reason \cite{socratic_cot}.

\section{\ourfull{}}

In this section, we define the objective of our proposed methodology \ourfull{} in detail. 
Given a query and an initial prediction generated by the LLM, \art{} uses a trainable pipeline for refinement as follows: (a) evaluate whether the initial generation requires refinement by asking a series of questions (\texttt{Ask}); (b) perform the \textit{refinement} step based on the evaluation (\texttt{Refine}); and finally (c) choose either the refined result or the initial prediction (\texttt{Trust}).  

\subsection{Initial Prediction}
Given a task query \texttt{x}, the LLM $\psi$ generates an initial prediction \texttt{y} $= \psi$ (\texttt{x}). For pre-trained LLMs, the query \texttt{x} is augmented with several examples of the task as few-shot prompts, while for fine-tuned models the query is provided directly without any examples. Due to the multi-step reasoning nature of the tasks where intermediate steps are beneficial for the model to arrive at the final answer, we consider using Chain of Thought \cite[CoT;][]{cot} and Subquestion Decomposition \cite[Decomp;][]{socratic_cot, leasttomost} as two of our main methods for initial prediction.

\subsection{\texttt{Ask}}
Once the initial prediction is generated, the next step is to decide when to refine the output. Refining every sample often leads to much worse performance \cite{Huang2023LargeLM}. Thus, we train an \asker{} to determine whether a prediction is correct or not, and then refine only the samples about which the \asker{} is uncertain about. However, before a smaller model can determine whether a generated answer is correct or whether refinement is needed, it is important to align the model with task-specific knowledge and the expected outcome. We fine-tune the smaller model in CoT style (intermediate steps with the final answer, as shown by the ``Initial Prediction'' in \autoref{fig:main:large}) on the training data. Next, we create the dataset for training the \asker{} model. We use the LLM $\psi$ to generate $k$ predictions per example on the training set, and then label them ``Yes'' or ``No'' for refinement based on whether the prediction was correct or incorrect (exact numbers are presented in \autoref{tab:data}). For each prediction, we append the subquestions present in the datasets \footnote{Note that the subquestions are available for each data set and we used them to train the \asker{} model. However, LLMs can be used to generate the subquestions and then distill them into smaller models that perform similarly to the ground truth \cite{magister-etal-2023-teaching}} prior to the ``Yes'' or ``No'' decision to further train the fine-tuned model. In this way the \asker{} learns to first ask the relevant questions, map them to the prediction and then decide whether all its questions are answered in the prediction or not, leading to the refinement decision. An example is presented in the appendix \autoref{fig:pipeline}.

\begin{table} [h]
\centering
\small
\begin{tabular}{ c | c c c}
    \toprule 
    & \multicolumn{3}{c}{Train Samples} \\
     \bf{Dataset} & \bf{Fine-tune} & \bf{\asker{}} & \bf{\truster{}}\\
    \midrule
    GSM8K & 7473 & 35000 & 15000 \\ 
    StrategyQA & 1832 & 9000 & 2300 \\
    \bottomrule
\end{tabular}
\caption{Comparison of different data sizes used for fine-tuning, and training the \asker{} and \truster{} models.}
\label{tab:data}
\end{table}

\subsection{\texttt{Refine}}
If the \asker{} predicts ``Yes'' (refinement is needed), then the LLM $\psi$ is used to refine the output given the input and  the subquestions from the \asker{} model, $\texttt{y}_{\text{ref}}$ = $\psi$(\texttt{x;subq}). Similar to \citet{Shridhar2023SCREWSAM}, for the StrategyQA dataset, additional facts (\texttt{facts}) are also provided to the model $\psi$ during refinement ($\texttt{y}_{\text{ref}}$ = $\psi$(\texttt{x;subq;facts})).  An example is presented in appendix \autoref{fig:ques-sample}. 

\subsection{\texttt{Trust}}
Finally, to decide whether the refinement output should be preferred over the original generation, we train a \truster{} that takes two candidates ($\texttt{y}$, $\texttt{y}_{\text{ref}}$) for the task query \texttt{x} and decides which one to prefer over the other. An example is presented in the appendix \autoref{sec:art-pipeline}. 
However, in 80\% of the cases, the final answer of the refinement $\texttt{y}_{\text{ref}}$ and the initial prediction $\texttt{y}$ were the same. Our goal is to make \truster{} learn to identify the reasoning chain with the correct final answer and not a particular styled intermediate reasoning chain. To create a good-sized training data set, we used the same training data collected for the \asker{} model (\autoref{tab:data}) and selected the prediction samples that have both a correct and an incorrect prediction. We construct preferred (correct predictions) vs. non-preferred (incorrect predictions) pairs and train a \truster{} with the text classification objective as:

\begin{equation}
    \mathcal{L}_\theta = -\mathbb{E}_{\texttt{x},y_j,y_k \sim \mathcal{D}}\ [\text{log}(\sigma (r_\theta(\texttt{x},y_j) - r_\theta(\texttt{x},y_k)))]
\end{equation}

where, $r$ is the score of the \truster{} model, $y_j$ is the preferred candidate (correct prediction) and $y_k$ is the non-preferred candidate (incorrect prediction in our case) from the dataset $\mathcal{D}$. 
Based on the score for each sample, we select the best scored output.

\section{Experiments}

\subsection{Dataset}

We test the \art{} refinement strategy on two multi-step reasoning tasks, GSM8K \citep{GSM8k} and StrategyQA \cite{strategyqa}. The GSM8K dataset is a grade school math word problem dataset with a training set of 7473 samples and a test set of 1319 samples, each requiring two to eight steps to solve. The dataset also consists of sub-questions that correspond to the steps in a given correct solution. StrategyQA, on the other hand, is a question-answering benchmark that focuses on open-domain questions, requiring reasoning steps to solve it.
StrategyQA consists of 2290 training examples, of which the first 20\% were used as the test set and the remaining 80\% as the training set, following previous work \cite{magister-etal-2023-teaching, Shridhar2023SCREWSAM}. Each question is accompanied by its decomposed questions and the correct factual knowledge required to answer it. Example of each dataset is presented in appendix \autoref{fig:ques-sample}. 

\subsection{Experimental Setup}
We use LLaMA 70B (pre-trained and chat) \cite{LLaMA2}, ChatGPT (turbo (\texttt{gpt-3.5-turbo}) and turbo-instruct (\texttt{gpt-3.5-turbo-instruct})). \cite{brown2020language}, and GPT-4 (\texttt{gpt-4}) \cite{openai2023gpt4} as base models $\psi$ due to their popularity and state-of-the-art performance. Next, we fine-tuned variants of the LLaMA model (7B, 13B, and 70B) on the GSM8K dataset and 7B and 13B on the StrategyQA dataset. All fine-tuned variants were further trained to get the \asker{} model to ask relevant questions and decide when to refine. Finally, we fine-tuned the LLaMA 13B  model to get the \truster{} that decides between the original and refined output. 
All pre-trained and fine-tuned LLaMA models were used with greedy decoding during testing  (temperature = 0 and top p = 1). To collect data for training, different temperatures were used to collect diverse samples (temperature = \{0, 0.3, 0.4, 0.7, 0.8\}) and $k$ was set to 5 to generate 5 predictions on the train set. All training was done on a cluster of 8 A100s 80GB each GPUs (except for the LLaMA 70B fine-tuning, which required 4 clusters of 8 A100s each).

\subsection{Results}

\begin{table*} [h!]
\centering
\small
\begin{tabular}{c | c c | c c c | c c}
    \toprule 
     \multicolumn{1}{c}{Model}  & \multicolumn{2}{c}{Initial Prediction} & \multicolumn{3}{c}{Refinement} & \multicolumn{2}{c}{Trust}\\
     \bf{Type} & \bf{Method} & \bf{Accuracy} & \bf{Subquestions} & \bf{Model} & \bf{Accuracy} &  \bf{Model} & \bf{Accuracy}\\
     \midrule
    \multicolumn{8}{c}{LLaMA 70B}\\
    \midrule
    \rowcolor{Blue}
    Pre-trained & CoT & 59.74 & No & Self &  59.07 & Self & 59.83 \\ 
    \rowcolor{Blue}
    Pre-trained & CoT & 59.74 & Yes & Self &  \underline{59.83} & Self & \underline{60.43} \\
    \rowcolor{Blue}
    Pre-trained & Decomp & 54.55 & No & Self &  55.11 & Self & 55.34 \\
    \rowcolor{Blue}
    Pre-trained & Decomp & 54.66 & Yes & Self &  50.26 & Self & 54.51 \\
    \rowcolor{Green}
    Pre-trained & CoT & 59.74 & Yes & \asker{}$_{\text{7B}}$ &  61.33 & \truster{} & 61.94 \\
    \rowcolor{Green}
    Pre-trained &CoT & 59.74 & Yes & \asker{}$_{\text{13B}}$ &  62.74 & \truster{} & 63.85  \\
    \rowcolor{Green}
    Pre-trained & CoT & 59.74 & Yes & \asker{}$_{\text{70B}}$ &  \textbf{63.60} & \truster{} & \textbf{64.24} \\
    \rowcolor{Blue}
    Chat & CoT & 58.90 & No & Self &  \underline{59.10} & Self & 58.79 \\ 
    \rowcolor{Blue}
    Chat & CoT & 58.90 & Yes & Self &  58.83 & Self & \underline{59.55} \\
    \rowcolor{Green}
    Chat & CoT & 58.90 & Yes & \asker{}$_{\text{7B}}$ &  60.12 & \truster{} & 61.18  \\
    \rowcolor{Green}
    Chat &CoT & 58.90 & Yes & \asker{}$_{\text{13B}}$ &  63.00 & \truster{} & 63.30  \\
    \rowcolor{Green}
    Chat & CoT & 58.90 & Yes & \asker{}$_{\text{70B}}$ &  \textbf{63.80} & \truster{} & \textbf{64.40} \\
    \midrule
    \multicolumn{8}{c}{ChatGPT}\\
    \midrule 
    \rowcolor{Yellow}
    Turbo& $\text{CoT}^S$ & 71.64 & $\text{No}^S$ & $\text{Self}^S$ & 73.00 & $\text{Self}^S$ & 72.93 \\ 
    \rowcolor{Yellow}
    Turbo & $\text{CoT}^S$ & 71.64 & $\text{Yes}^S$ & $\text{Self}^S$ & 73.99 & $\text{Self}^S$ & 73.99 \\ 
    \rowcolor{Yellow}
    Turbo & $\text{CoT}^{SR}$ & 74.58 & $\text{No}^{SR}$ & $\text{Self}^{SR}$ & 75.00 & $\text{Most Recent}^{SR}$ & 75.00 \\
    \rowcolor{Yellow}
    Turbo & $\text{CoT}^{SR}$ & 74.58 & $\text{No}^{SR}$ & $\text{Self}^{SR}$ & 75.00 & $\text{Most Recent}^{SR}$ & 75.00 \\
    \rowcolor{Yellow}
    Turbo & $\text{CoT}^{!C}$ & 75.90 & $\text{No}^{!C}$ & $\text{Self}^{!C}$ & 75.10 & $\text{Most Recent}^{!C}$ & \underline{75.10} \\
    \rowcolor{Blue}
    Turbo & CoT & 77.71 & No & Self & 78.16 & Self & 78.28 \\
    \rowcolor{Blue}
    Turbo & CoT & 77.71 & Yes & Self & 78.46  & Self & 78.89  \\
    \rowcolor{Blue}
    Turbo & Decomp & 78.62 & No & Self & \underline{78.99}  & Self & 78.99  \\
    \rowcolor{Blue}
    Turbo & Decomp & 78.62 & Yes & Self & 78.24  & Self & \underline{79.22}  \\
    \rowcolor{Green}
    Turbo & CoT & 77.71 & No & \asker{}$_{\text{7B}}$ & 80.89 & \truster{} &  81.14\\
    \rowcolor{Green}
    Turbo & CoT & 77.71 & Yes & \asker{}$_{\text{13B}}$ & \bf{82.18} & \truster{} &  \bf{82.64}\\
    \rowcolor{Blue}
    Instruct & CoT & 71.26 & No & Self & 70.28 & Self & 71.50 \\
    \rowcolor{Blue}
    Instruct & CoT & 71.26 & Yes & Self & \underline{72.32} & Self & 72.85 \\
    \rowcolor{Green}
    Instruct & CoT & 71.26 & Yes & \asker{}$_{\text{7B}}$ & 76.19 & \truster{} & 76.34 \\ 
    \rowcolor{Green}
    Instruct & CoT & 71.26 & Yes & \asker{}$_{\text{13B}}$ &  \bf{78.46} & \truster{} & \bf{79.86} \\

    \midrule
    \multicolumn{8}{c}{GPT-4}\\
    \midrule
    \rowcolor{Yellow}
    - & $\text{CoT}^S$ & 91.45 & $\text{Yes}^S$ & $\text{Self}^S$ & 90.80 & $\text{Self}^S$ & \underline{93.10} \\ 
    \rowcolor{Yellow}
    - & $\text{CoT}^{SR}$ & 92.90 & $\text{No}^{SR}$ & $\text{Self}^{SR}$ & \underline{93.10} & $\text{Most Recent}^{SR}$ & \underline{93.10} \\
    \rowcolor{Green}
    - & \text{CoT} & 91.88 & Yes & \asker{}$_{\text{7B}}$ & 93.25 &  \truster{} &  93.45\\
    \rowcolor{Green}
    - & \text{CoT} & 91.88 & Yes & \asker{}$_{\text{13B}}$ & \bf{93.72} &  \truster{} &  \textbf{94.08}\\
    \bottomrule
\end{tabular}
\caption{Accuracy (\texttt{maj1@1}) comparison between different methods and refinement strategies on the GSM8K dataset. \texttt{Initial Prediction} refers to the initial generation from the LLM with its \texttt{Method} referring to one of the reasoning strategies (Chain of Thought (CoT) or Subquestion Decomposition (Decomp) in our case). \texttt{Refinement} refers to the combination of the \emph{Ask} and the \textit{Refine} stages in \art{} with or without the use of subquestions during refinement (\texttt{subquestions}). Finally, \texttt{Trust} refers to the \emph{Trust} stage in \art{}, where \emph{Self} refers to \emph{self-refinement}, \truster{} is our trained model and \emph{Most Recent} refers to choosing refinement as the final result. \colorbox{Yellow}{Yellow} represents the baseline methods from previous work ((.)$^S$ represents results from \citet{Shridhar2023SCREWSAM}, (.)$^{SR}$ from \citet{SelfRefine}, and (.)$^{!C}$ from \citet{Huang2023LargeLM}), \colorbox{Blue}{Blue} represents our implementations of the baselines, and \colorbox{Green}{Green} represents our proposed methods.  \underline{Underline} represents the best results from previous strategies, and \textbf{bold} represents the overall best result.}
\label{tab:main}
\end{table*}

\paragraph{Self-Refinement is not enough}
\label{sec:establishing}
\autoref{tab:main} shows the refinement framework of initial prediction, refinement, and trust. In general, the performance of LLaMA 70B is much lower than the ChatGPT turbo model for the GSM8K dataset (59 compared to 77 for CoT and 55 compared to 78 for Subquestion Decomposition). Furthermore, the Subquestion Decomposition (Decomp) approach performs better than CoT for ChatGPT, but the opposite is true for LLaMA 70B. Since the training data and the model architecture of ChatGPT are not public, it is difficult to understand the performance gap. Finally, \emph{self-refinement} improves performance in some cases, but leads to worse performance in others (\colorbox{Blue}{Blue} colored boxes in \autoref{tab:main} show the comparison). However, combining refinement with the trust module consistently improves performance over the initial prediction in almost all cases. This demonstrates the usefulness of the different components of our proposed \art{} methodology . Note that our baselines of the \texttt{Self} modules of refinement and trust uses the same prompts as presented in \citet{Shridhar2023SCREWSAM} for a fair comparison.

\paragraph{Importance of \texttt{Asking}} \autoref{tab:main} demonstrates the effectiveness of training an \asker{} that decides when to refine the outputs. Compared to the self-refinement (\texttt{Self}) strategy, a much smaller model like LLaMA 7B (\asker{}$_{\text{7B}}$) outperforms much larger LLMs like ChatGPT self-refinement (\texttt{Self}) by over 2 points (80.89 vs. 78.62). LLaMA 13B (\asker{}$_{\text{13B}}$) improves it by over 4 points (78.62 $\rightarrow$ 82.18). The trend is similar when refinements are compared with the self-refinement   capabilities (\texttt{Self}) of LLaMA 70B, where a 7B model (\asker{}$_{\text{7B}}$) outperforms the pre-trained self-refinement capabilities of LLaMA 70B by about 2 points (61.33 vs. 59.83) and over 1 point for the chat model (58.83 vs. 60.12). The 13B model (\asker{}$_{\text{13B}}$), on the other hand, improves it by over 3 points for the pretrained LLaMA 70B model (59.83 $\rightarrow$ 62.74) and the chat version by more than 4 points (58.83 $\rightarrow$ 63.00). Finally, using the 70B model as \asker{} (\asker{}$_{\text{70B}}$) further improves the results by 4 points for the pre-trained version (59.83 $\rightarrow$ 63.60) and over 5 points for the chat version (58.83 $\rightarrow$ 63.80).  The results follow a similar trend for the GPT-4 models, where both the 7B (\asker{}$_{\text{7B}}$) and 13B (\asker{}$_{\text{13B}}$) models improve the results over the initial generation by about 2 points (91.88 $\rightarrow$ 93.72), which is higher than other baselines from \citet{SelfRefine} and \citet{Shridhar2023SCREWSAM}. Finally, note that our proposed strategy \art{} improves the overall performance of ChatGPT to 82.18 after refining with a single pass (\texttt{maj1@1}), which is similar to the self-consistency score of 3 samples (\texttt{maj1@3}) \cite{Huang2023LargeLM}.

The results on StrategyQA follow a similar trend, where a 7B model \asker{}$_{\text{7B}}$ improves the refinement score by 1 point for LLaMA 70B (75.15 $\rightarrow$ 76.22) and over 3 points for ChatGPT (70.52 $\rightarrow$ 73.84), as shown in \autoref{tab:sqa}. Note that following \citet{Shridhar2023SCREWSAM}, we also provide some factual information along with the questions during refinement so that the model can correct its factual inaccuracy. The gains are larger for the \asker{}$_{\text{13B}}$ model, where the performance improves by 3 points for LLaMA 70B (75.15 $\rightarrow$ 78.38) and 5 points for ChatGPT (70.52 $\rightarrow$ 75.76), demonstrating the clear importance of asking questions for refinement decision making. 

\begin{table} [h]
\centering
\small
\begin{tabular}{ c  | c c | c c}
    \toprule 
     \multicolumn{1}{c}{Initial Pred} & \multicolumn{2}{c}{Refinement} & \multicolumn{2}{c}{Trust} \\
     \bf{Acc} & \bf{Model} & \bf{Acc} &  \bf{Model} &
 \bf{Acc}\\
    \midrule
    \multicolumn{5}{c}{LLaMA 70B Pre-trained}\\
    \midrule
    \rowcolor{Blue}
    74.45 & Self & 75.15 & Self &  75.74 \\
    \rowcolor{Green}
    74.45 & \asker{}$_{\text{7B}}$ & 76.22 & \truster{} & 76.12\\
    \rowcolor{Green}
    74.45 & \asker{}$_{\text{13B}}$ & \bf{78.38} & \truster{} & \bf{78.44}\\
    \midrule
    \multicolumn{5}{c}{ChatGPT Turbo}\\
    \midrule
    \rowcolor{Blue}
    73.58 & Self & 70.52 & Self & 74.89 \\
    \rowcolor{Green}
    73.58 & \asker{}$_{\text{7B}}$ & 73.84 & \truster{} & 74.04\\
    \rowcolor{Green}
    73.58 & \asker{}$_{\text{13B}}$ & \bf{75.76} & \truster{} & \bf{75.86}\\
    \bottomrule
\end{tabular}
\caption{Accuracy comparison on the StrategyQA dataset for refinement and trust with different models. \colorbox{Blue}{Blue} represents our implementations of the baselines, and \colorbox{Green}{Green} represents our proposed methods.}
\label{tab:sqa}
\end{table}

\paragraph{(Don't) Always Trust Refinement} \autoref{tab:main} demonstrates the usefulness of a trust module that decides whether the refinement improves or degrades the initial prediction. We train a \truster{} model that learns to rank the initial prediction and the refined output and decides which one to choose for a given input. Our trained \truster{} model (LLaMA 13B) achieves an accuracy of the pre-trained LLaMA 70B of as high as 64.24, which is 4 points higher than the baseline (60.43). The trend is similar for the chat version, where the improvement is almost 5 points over the baseline method of using the same LLM for decision making (59.55 $\rightarrow$ 64.40). The results follow a similar trend for ChatGPT where the improvement over baselines (the same LLM) is about 4 points for the Turbo model over the baselines (78.89 $\rightarrow$ 82.64) and about 7 points from the best previous method of Self-Refine \cite{SelfRefine} (75.10 of Self-Refine $\rightarrow$ 82.64). The gains for GPT-4 are very small, possibly due to the high performance of the GPT-4 model, but \truster{} improves the performance to 94.08 from the previous best refinement score of 93.10.

For StrategyQA, the trust module does not prove to be very helpful with a performance very similar to the refinement scores. This shows that it is difficult to train a \truster{} on fact-based datasets, as it is hard to rank two pieces of factual information without knowing the true facts.

\paragraph{Cost of fine-tuning LLMs vs. \art{}-based refinement}
Since the training samples are available for the GSM8K dataset, it is possible to fine-tune a LLaMA 70B model. Fine-tuning LLaMA 70B achieves 63.2\% accuracy on GSM8K \cite{Yuan2023ScalingRO}, which is similar to what a trained 13B \asker{}$_\text{13B}$ and \truster{} can achieve with a pre-trained LLaMA 70B model, while incurring much lower training costs and computational requirements. \autoref{tab:flops} shows that training a 13B model as \truster{} is 10X cheaper than fine-tuning a 70B model, and even with two trained models as \asker{} and \truster{}, \art{} is still 5X cheaper. 
In addition, fine-tuning usually makes the model narrowly specialized to the trained dataset with reduced general in-context learning performance \cite{Wang2022TwostageLF}, which won't happen with a pre-trained model deciding when to refine using our proposed framework \art{}.

\begin{table} [h]
\centering
\small
\begin{tabular}{ c c  c  c }
    \toprule 
     \bf{Objective} & \bf{Model Size} & \bf{Flops} & \bf{GPU Hours}\\
    \midrule
    \asker{} & 7B & 1.5 X 10$^{17}$ & 1 \\ 
    \truster{} & 13B & 3 X 10$^{17}$ & 4 \\
    FineTuning & 70B &  1.5 X 10$^{18}$ & 75\\
    \bottomrule
\end{tabular}
\caption{Comparison of different compute requirements for training different sized LLaMA models on GSM8K with the objective of training a decision maker (\asker{} and \truster{}) vs. finetuning a model (FineTuning).}
\label{tab:flops}
\end{table}

\section{Ablation Studies}

\paragraph{Importance of Asking Questions for Refinement}
We trained \asker{} to make only a binary decision of ``Yes'' or ``No'' to refine, without asking the relevant questions, and found that all versions of the LLaMA models always trusted the predictions and never decided to refine them. LLMs are often very bad at judging their own predictions and often prefer their own predictions \cite{kadavath2022language}, and our experiments observed a similar phenomenon. However, asking questions leads to a better refinement decision and a qualitative example is presented in \autoref{fig:qualitative}.

\begin{figure} [h!]
    \centering
        \includegraphics[width=0.45\textwidth]{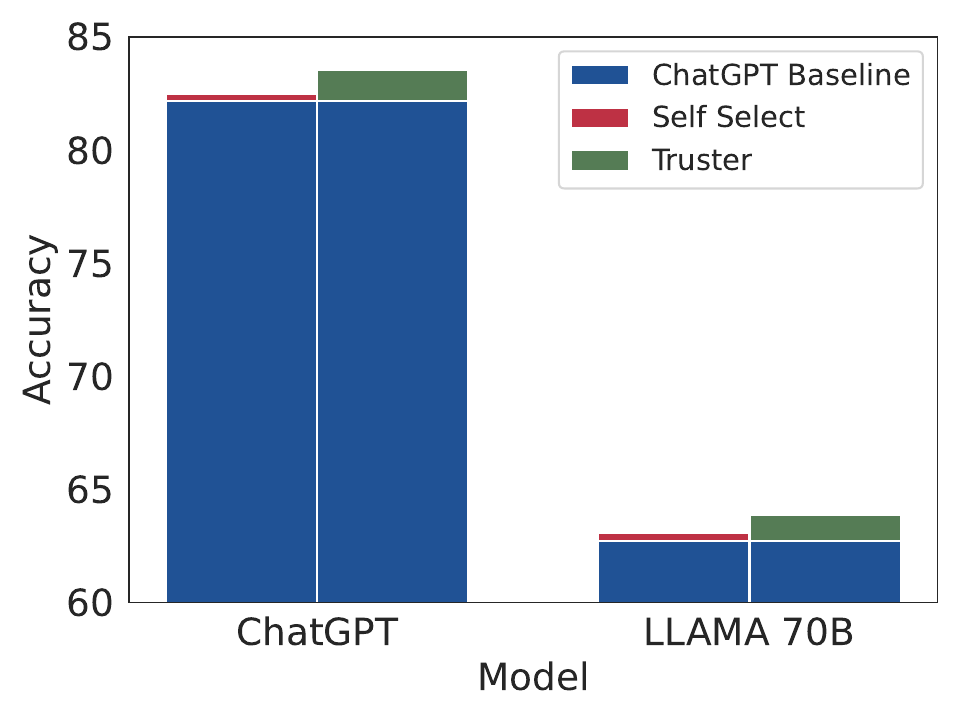}
    \caption{Comparison of the trained \truster{} with the self-selection version of the LLMs for GSM8K.}
    \label{fig:ranker}
\end{figure}

\paragraph{Importance of \truster{} for selection}
We compared the performance of the selection module of the LLM (Self) vs. our trained \truster{} for the GSM8K dataset and observed that the trained \truster{} can better assess the errors made in the predictions and asks the model to revert to the previous generation more (about 50\% more compared to self-selection); leading to superior performance (\autoref{fig:ranker}).

\begin{figure}
    \centering
        \includegraphics[width=0.45\textwidth]{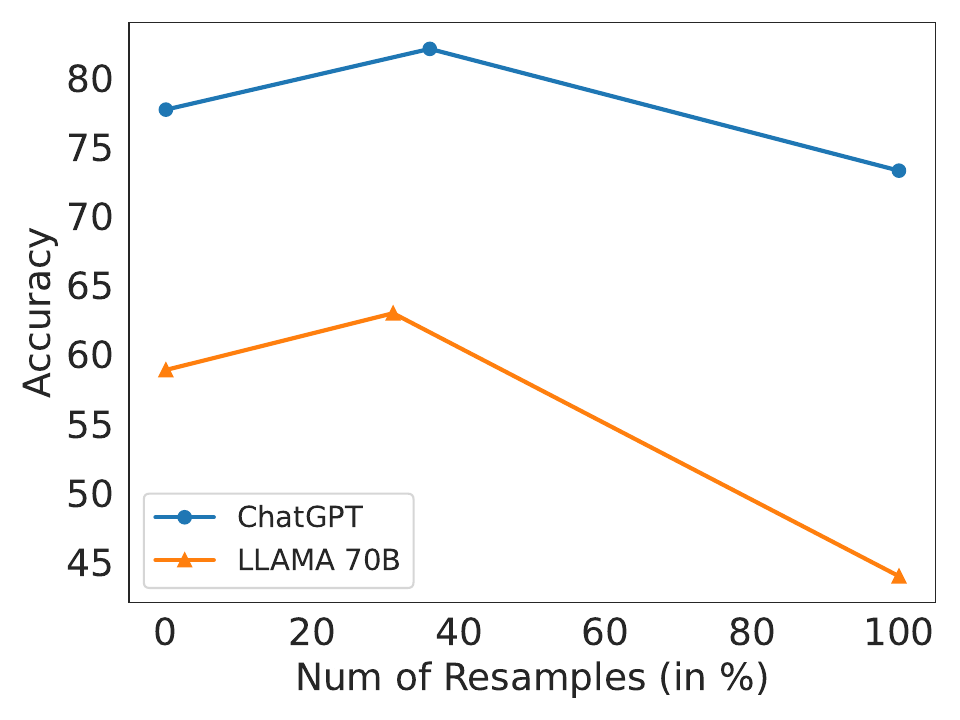}
    \caption{Number of resamples for refinement for ChatGPT and LLaMA 70B models on GSM8K. 0 means no resampling and 100 means resampling all the time.}
    \label{fig:resample}
\end{figure}

\paragraph{When to refine?}
Assessing when to refine is an important component of the refinement pipeline, as always refining leads to worse results \cite{Huang2023LargeLM}. \autoref{fig:resample} supports the previous findings and shows that always refining can hurt the overall performance (100\% refinement) and is worse than the initial prediction (0\% refinement). The sweet spot is somewhere in the middle (about 30-35\% refinement seems to work for both ChatGPT and LLaMA 70B models on the GSM8K dataset).

\paragraph{Can \asker{} be trained on its own output?}
Instead of training an \asker{} to ask questions on the output of the pre-trained LLM, can we train it on its own predictions? Much to our surprise, we find that \asker{} trained on its own data can make a better refinement decision than LLM's self-refinement. However, as expected, training on pre-trained model data proved to be more beneficial in deciding when to refine, due to a similar distribution of test and training samples, as shown in \autoref{tab:selfvslarge}. However, for ChatGPT models, \asker{} trained on its own data performs similarly to that trained on LLaMA 70B models, as both data distributions are different from the ChatGPT test distribution (82.10 vs. 82.18 for the 13B model and 80.69 vs. 80.89 for the 7B model). 

\begin{table} [h]
\centering
\small
\begin{tabular}{ c  | c c c }
    \toprule 
     \multicolumn{1}{c}{Initial Prediction} & \multicolumn{3}{c}{Refinement}\\
     \bf{Accuracy} & \bf{Model} & \bf{Data} & \bf{Accuracy}\\
    \midrule
    \multicolumn{4}{c}{LLaMA 70B Pre-trained}\\
    \midrule
    59.74 & \asker{}$_{\text{7B}}$ & 7B &  59.21\\ 
    59.74 & \asker{}$_{\text{7B}}$ & 70B &  \bf{61.33}\\
    59.74 & \asker{}$_{\text{13B}}$ & 13B &  62.39\\ 
    59.74 & \asker{}$_{\text{13B}}$ & 70B &  \bf{62.74}\\ 
    \midrule
    \multicolumn{4}{c}{ChatGPT Turbo}\\
    \midrule
    77.71 & \asker{}$_{\text{7B}}$ & 7B & 80.69 \\ 
    77.71 & \asker{}$_{\text{7B}}$ & 70B &  \bf{80.89}\\ 
    77.71 & \asker{}$_{\text{13B}}$ & 13B & 82.10\\ 
    77.71 & \asker{}$_{\text{13B}}$ & 70B & \bf{82.18}\\

    \bottomrule
\end{tabular}
\caption{Comparison of refinement results when \asker{} models are trained on their own data vs. LLaMA 70B pre-trained model data on the GSM8K dataset.}
\label{tab:selfvslarge}
\end{table}

\begin{figure*}[t]
\small
\centering
\begin{tabular}{l}
\toprule
\addlinespace
\parbox{\textwidth}{\textbf{Question} A customer’s loyalty card at a store gives them rewards of \$1 off their next purchase for every \$20 they spend. Their last shopping trip, they spent \$80. This shopping trip, they spent \$43, used their rewards, and applied a coupon that took twice the amount of rewards off the price. How many dollars did the customer pay on this shopping trip?} \\
\addlinespace
\parbox{\textwidth}{\textbf{Initial Prediction} The customer spent \$80 on their last shopping trip, so they earned 80/20 = <<80/20=4>>4 rewards. They used 2*4 = <<2*4=8>>8 rewards on this shopping trip. So the price of their purchase was \$43 - \$8 = \$<<43-8=35>>35. The customer paid \$35 on this shopping trip. The answer is 35. \textcolor{red}{\ding{55}}} \\
\addlinespace
\parbox{\textwidth}{\textbf{Self-Refinement} Yes, the reasoning is correct. The customer spent \$80 on their last shopping trip, earning them 80/20 = 4 rewards. They used 8 rewards on this shopping trip.
Therefore, the price of their purchase was \$43 - \$8 = \$35. \textcolor{red}{\ding{55}}} \\
\addlinespace
\parbox{\textwidth}{\textbf{\asker{}} How much did the customer get in rewards? What was the coupons cost? How much did the customer pay on this shopping trip? All questions are not answered.} \\
\addlinespace
\parbox{\textwidth}{\textbf{Refinement}
\colorbox{Green}{How much did the customer get in rewards?} The customer had 80 / 20 = \$<<80/20=4>>4 in rewards. \\ \colorbox{Green}{What was the coupons cost?} Their coupon was worth 4 * 2 = \$<<4*2=8>>8.  \colorbox{Green}{How much did the customer pay on this shopping trip?} Thus they paid 43 - 4 - 8 = \$<<43-4-8=31>>31 on this shopping trip. \textcolor{green}{\ding{51}}} \\

\addlinespace
\bottomrule
\end{tabular}
\caption{Qualitative example demonstrating the usefulness of asking questions to decide \emph{when to refine} and then using \colorbox{Green}{those questions} to refine the initial prediction by fixing the errors.}
\label{fig:qualitative}
\end{figure*}

\paragraph{Can \art{} methodology be extended to SoTA models?}
We fine-tune MetaMath 7B and 13B \cite{MetaMath} using the same training strategy as \asker{} models and compare the results in \autoref{tab:sota}. MetaMath, due to its training on additional data and superior mathematical reasoning abilities, can evaluate the prediction better and leads to a better overall performance after refinement (62.31 vs. 61.33 for 7B and 64.06 vs. 62.74 for 13B variant). MetaMath models also perform better with fewer samples (about 27\% fewer samples for 7B and 25\% fewer for 13B), suggesting that superior mathematical reasoning can help to better evaluate predictions, leading to fewer uncertain samples for refinement. Since MetaMath was trained on over 250K samples with rejection sampling, it was not possible for us to run all experiments on this large dataset, and we stuck to LLaMA models for all of our experiments.  

\begin{table} [h]
\centering
\small
\begin{tabular}{ c | c c c }
    \toprule 
     \multicolumn{1}{c}{Initial pred} & \multicolumn{3}{c}{Refinement}\\
     \bf{Accuracy} & \bf{\asker{} } & \bf{Acc ($\uparrow$)} & \bf{\% samp ($\downarrow$)}\\
    \midrule
    
    59.74 & LLaMA 7B & 61.33 & 48\\ 
    59.74 & MetaMath 7B & \bf{62.31} & 35 \\
    59.74 & LLaMA 13B & 62.74 & 36 \\ 
    59.74 & MetaMath 13B &  \bf{64.06} & 27\\ 
    \bottomrule
\end{tabular}
\caption{Comparison of LLaMA 7B and 13B refinement accuracy (Acc) with the state-of-the-art MetaMath 7B and 13B models \cite{Yuan2023ScalingRO} and their sampling percentage (\% samp) for refinement.}
\label{tab:sota}
\end{table}

\begin{figure}
    \centering
        \includegraphics[width=0.5\textwidth]{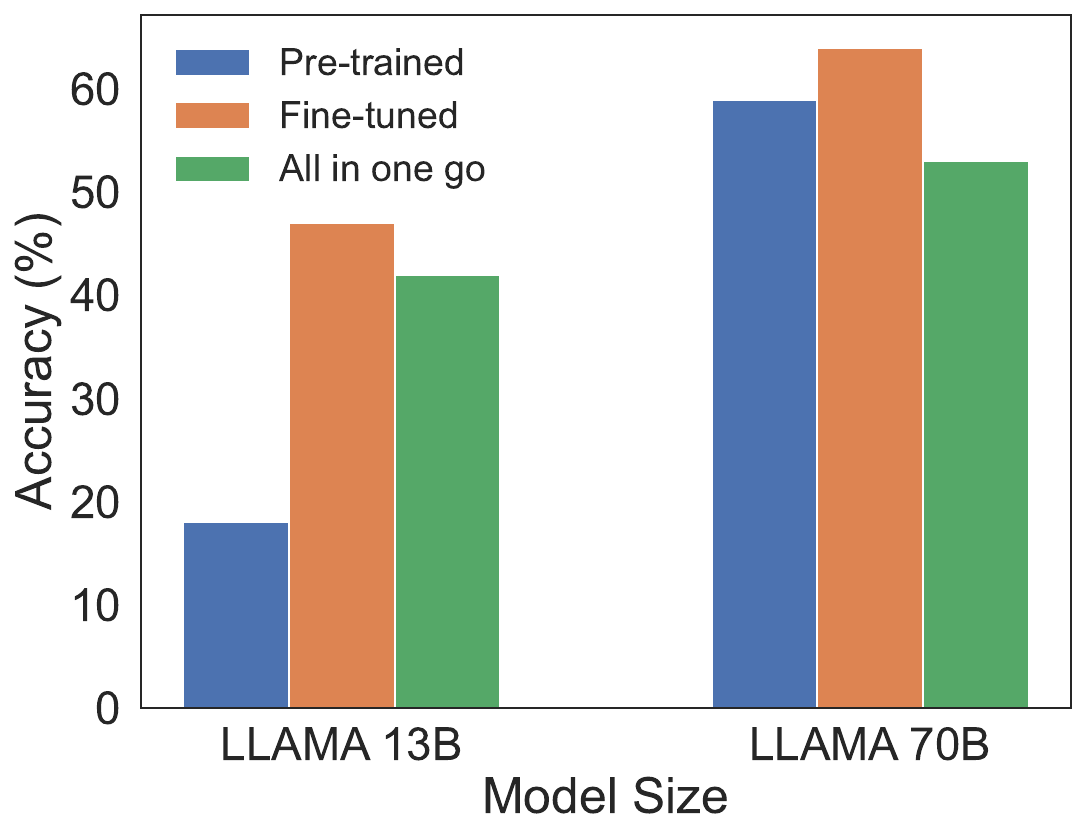}
    \caption{Comparison of the all-in-one approach to fine-tuning the LLMs on the GSM8K dataset.}
    \label{fig:all-in-one-go}
\end{figure}

\paragraph{Entire \art{} pipeline in one go} 
To test whether the entire \art{} pipeline of asking relevant questions, then deciding whether the questions are answered or not, and then refining can be learned in one go instead of individual models for each step, we train a LLaMA 13B and 70B model over the entire sequence (all-in-one-go).  \autoref{fig:all-in-one-go} shows that all-in-one-go (green) performs worse than fine-tuning (orange) for the LLM, demonstrating that generating the entire sequence is a more challenging task for the LLM than individual components.

\section{Key Findings}
From the experiments, we observe the following:
\begin{itemize}
    \item \textbf{\art{} allows smaller models to make refinement decisions superior to  LLM self-refinement}: Smaller models trained to make a refinement decision can outperform a much larger model in \emph{self-refinement} style (\autoref{tab:main}).
    \item \textbf{Ask questions before refinement}  Asking questions is an effective way to verify the quality of the generations and allows the models to make better refinement decisions.
    \item \textbf{Smaller models' refinement decisions are a cost-effective alternative to fine-tuning LLMs} The refinement decision of smaller models combined with a pre-trained LLM performs similarly to a larger model when fine-tuned. This saves a lot of computation required to fine-tune a larger model (\autoref{tab:flops}) and preserves downstream performance on other tasks. 
    \item \textbf{Expert models can make better judgments about refinement} Larger models (\asker{}$_\text{13B}$ performance is better than \asker{}$_\text{7B}$ in all cases) show that better models can make more informed decisions about when to refine. \autoref{tab:sota} shows that MetaMath trained models outperform LLaMA models of similar size. 
    \item \textbf{Trained \truster{} can rank decisions better} A trained smaller \truster{} model can rank the results better than the self-selection version of LLMs, as shown in \autoref{fig:ranker}.
\end{itemize}

\section{Conclusion}
In this work, we propose a refinement strategy called \ourfull{}, which allows smaller models to make refinement decisions for LLMs and determine whether these refinements are reliable. We empirically demonstrate the effectiveness of our approach on two reasoning tasks, mathematical word problems and question answering. Our results show that smaller models, even up to 10X smaller, can outperform larger models in making refinement decisions.

\section*{Limitations}
In this work, we trained a \asker{} to make a refinement decision by asking questions to verify the predictions. We used the training data available for the GSM8K and StrategyQA datasets. However, for many tasks, training data may not be available. In such cases, LLMs can be used to generate data and in many cases it performs similarly to the ground truth data \cite{magister-etal-2023-teaching}. However, we have not tested this with \art{} due to the availability of the training dataset. In addition, for StrategyQA, we used the available facts to support the model decision when refining the predictions. These facts were available in the dataset, but in the real world can be extracted with the help of some tools or from some databases. We did not test this approach in our work and leave it for future work.

\newpage
\bibliography{anthology,custom}

\newpage
\appendix

\label{sec:art-pipeline}
\begin{figure*}[t]
\small
\centering
\begin{tabular}{l}
\toprule
\addlinespace
\parbox{\textwidth}{\textbf{Question} A customer’s loyalty card at a store gives them rewards of \$1 off their next purchase for every \$20 they spend. Their last shopping trip, they spent \$80. This shopping trip, they spent \$43, used their rewards, and applied a coupon that took twice the amount of rewards off the price. How many dollars did the customer pay on this shopping trip?} \\
\addlinespace
\parbox{\textwidth}{\textbf{Initial Prediction} The customer spent \$80 on their last shopping trip, so they earned 80/20 = <<80/20=4>>4 rewards. They used 2*4 = <<2*4=8>>8 rewards on this shopping trip. So the price of their purchase was \$43 - \$8 = \$<<43-8=35>>35. The customer paid \$35 on this shopping trip. The answer is 35. \textcolor{red}{\ding{55}}} \\
\addlinespace
\parbox{\textwidth}{\textbf{\texttt{Ask}} How much did the customer get in rewards? What was the coupons cost? How much did the customer pay on this shopping trip? All questions are not answered.} \\
\addlinespace
\parbox{\textwidth}{\textbf{\texttt{Refine}}
\emph{How much did the customer get in rewards?} The customer had 80 / 20 = \$<<80/20=4>>4 in rewards. \emph{What was the coupons cost?} Their coupon was worth 4 * 2 = \$<<4*2=8>>8. \emph{How much did the customer pay on this shopping trip?} Thus they paid 43 - 4 - 8 = \$<<43-4-8=31>>31 on this shopping trip. \textcolor{green}{\ding{51}}} \\
\addlinespace
\parbox{\textwidth}{\textbf{\texttt{Trust}}
The customer had 80 / 20 = \$<<80/20=4>>4 in rewards. Their coupon was worth 4 * 2 = \$<<4*2=8>>8. Thus they paid 43 - 4 - 8 = \$<<43-4-8=31>>31 on this shopping trip. \textcolor{green}{\ding{51}}} \\
\midrule
\addlinespace
\parbox{\textwidth}{\textbf{Question} The cell-phone recharges at the rate of 1 percentage-point of charge per 3 minutes. Now, the phone is at 60\% charged. How long will it take to fully charge, in hours?} \\
\addlinespace
\parbox{\textwidth}{\textbf{Initial Prediction} The phone needs 100-60 = <<100-60=40>>40\% more charge to be fully charged. Since the phone recharges at the rate of 1 percentage-point of charge per 3 minutes, it will take 40*3 = <<40*3=120>>120 minutes to fully charge. Therefore, it will take 120/60 = <<120/60=2>>2 hours to fully charge. The answer is 2. \textcolor{green}{\ding{51}}} \\
\addlinespace
\parbox{\textwidth}{\textbf{\texttt{Ask}} How many percentage points is left to be charged? How many minutes will it take to fully charge? How long will it take to fully charge, in hours? All questions are answered.} \\
\bottomrule
\end{tabular}
\caption{End to end flow of our proposed objective of \ourfull{} if \asker{} decides to refine (top) vs no refinement needed (bottom).}
\label{fig:pipeline}
\end{figure*}

\begin{figure*}[t]
\small
\centering
\begin{tabular}{l}
\toprule
\addlinespace
\parbox{\textwidth}{\textbf{Question} Claire makes a 3 egg omelet every morning for breakfast.  How many dozens of eggs will she eat in 4 weeks? } \\
\addlinespace
\parbox{\textwidth}{\textbf{Subquestions} How many eggs does Claire eat in a week? How many eggs will Claire eat in 4 weeks? How many dozens of eggs will Claire eat in 4 weeks? } \\
\addlinespace
\parbox{\textwidth}{\textbf{Answer} She eats 3 eggs every day and there are 7 days in a week so she eats 3*7 = <<3*7=21>>21 eggs a week. After 4 weeks she will have eaten 4*21 = <<4*21=84>>84 eggs. There are 12 eggs in 1 dozen and she'll eat 84 eggs so that's 84/12 = <<84/12=7>>7 dozen eggs. } \\
\addlinespace
\midrule
\addlinespace
\parbox{\textwidth}{\textbf{Question} Can a greyhound walk on two legs?} \\
\addlinespace
\parbox{\textwidth}{\textbf{Subquestions} What type of animal is a greyhound? Does \#1 walk on two legs? } \\
\addlinespace
\parbox{\textwidth}{\textbf{Facts} Greyhounds are dogs. Dogs walk on four legs. } \\
\addlinespace
\parbox{\textwidth}{\textbf{Answer} False} \\
\bottomrule
\end{tabular}
\caption{Example of a GSM8K data sample (top) and StrategyQA data sample (bottom).}
\label{fig:ques-sample}
\end{figure*}

\end{document}